\documentclass[runningheads]{llncs}
\usepackage[final,year=2026,ID=22]{accv}
\usepackage{accvabbrv}
\usepackage{graphicx}
\usepackage{booktabs}
\usepackage{multirow}
\usepackage{amsmath}
\usepackage[accsupp]{axessibility}
\usepackage[pagebackref,breaklinks,colorlinks,citecolor=accvblue]{hyperref}
\newcommand{\ci}[3]{\begin{tabular}[t]{@{}c@{}}#1\\[-2pt]{\scriptsize[#2,#3]}\end{tabular}}

\begin{document}
\title{Changing the Script Is Not Recovering the Answer: Donor-Content Controls for Multilingual OCR-VQA}
\titlerunning{Donor-Content Controls for Multilingual OCR-VQA}
\author{Surapon Nonesung\inst{1} \and
Pakin Siwathammarat\inst{2} \and
Natapong Nitarach\inst{1} \and
Warit Sirichotedumrong\inst{1} \and
Pittawat Taveekitworachai\inst{1} \and
Kunat Pipatanakul\inst{1}}
\authorrunning{S. Nonesung et al.}
\institute{Typhoon Team, SCB DataX, Bangkok, Thailand \and
Vidyasirimedhi Institute of Science and Technology (VISTEC), Rayong, Thailand}
\maketitle

\begin{abstract}
Multilingual vision--language models (VLMs) show large performance gaps in
OCR-based visual question answering, raising questions about why they fail on
languages such as Thai and Hindi. One possible explanation is language
commitment: models may hold useful answer information but fail to express it
in the target language. If so, shifting the model's language state might
recover some failed answers. Activation patching can alter output-language behavior by
replacing hidden states, but the English donor may also introduce answer
content, making apparent recovery difficult to interpret. We therefore ask whether hidden-state interventions recover
previously failed answers and how donor content affects that recovery. Using three frozen VLMs,
parallel rendered-XQuAD passages, and real Thai images, we examine output
scripts, LogitLens signals, aligned and shuffled donor patching, and
mean-difference steering with random-direction controls. Aligned patching
produces English-gold matches on 15--43\% of baseline Thai failures, but
shuffled donors reduce these matches to 4.4--5.3\% (none under exact matching)
while preserving substantial script changes. Held-out steering yields modest
recovery above random controls, ranging from 4.5 to 8.3 percentage points
across Thai and Hindi (1.0--6.0 under exact matching). Script control also
transfers to real Thai images without demonstrating improved answer accuracy.
These findings show that changing output language and recovering recipient
answer information are distinct effects, highlighting the need for
donor-content controls when interpreting hidden-state interventions.
\end{abstract}

\section{Introduction}
Multilingual OCR-VQA benchmarks reveal substantial differences across languages
\cite{pfeiffer-etal-2022-xgqa,changpinyo-etal-2023-maxm,tang-etal-2025-mtvqa,
nonesung-etal-2025-thaiocrbench}. In our rendered-XQuAD evaluation, Qwen3-VL-8B
scores 0.905 on English but 0.338 on Thai.

One possible explanation is what we call a failure of \emph{language
commitment}: the model may hold useful answer information but fail to express
it in the target language. Text-only studies make this plausible: models often
represent content in a largely language-shared space and can fail when
converting it into the target language
\cite{wendler-etal-2024-latent-language,wang-etal-2025-lost-multilinguality,bafna-etal-2025-translation}.
If so, shifting the model's language state might recover some failed answers.
Patching in an English state changes it, but the donor state can also carry the
answer, so apparent recovery may come from the donor rather than the model.
We first ask whether failures look like a language problem
(\S\ref{sec:diagnostics}). We then compare aligned with shuffled English donors,
whose state comes from a different item, and test steering, which copies no
single donor's state, against a random direction (\S\ref{sec:interventions}). Finally, we test real Thai images. We
distinguish \emph{script change}, \emph{recipient-answer
matching}, and \emph{donor-content dependence}: script changes survive donor
mismatch, whereas English-gold recovery falls sharply.

\section{Related Work}
Multilingual VQA and OCR benchmarks characterize performance across languages
and document types \cite{geigle-etal-2024-babel,tang-etal-2025-mtvqa,
jiu-etal-2025-tvqacml,nonesung-etal-2025-thaiocrbench}. We use parallel
XQuAD \cite{artetxe-etal-2020-cross} to align passage and question content.

VLM analyses show that useful visual representations need not be used
successfully by generation \cite{fu-etal-2025-hidden,neo-etal-2025-visual-processing}.
Text-only studies identify multilingual representations and output-language
effects \cite{wendler-etal-2024-latent-language,
marchisio-etal-2024-language-confusion,cocoa-2025-language-adherence},
and VLMs show related language priors \cite{long-etal-2026-language-prior}.
Hinck \etal\ \cite{liu-etal-2024-llava-english} ask why LLaVA models reply in
English; they study \emph{which} language a model answers in, whereas we ask
whether changing that language recovers the answer. Dumas \etal\
\cite{dumas-etal-2025-separating} use activation patching to separate
language and concept in translation. Our novelty is not the general
observation that late states affect output language. We examine how donor
content confounds an answer-recovery interpretation of patching in
image-conditioned QA, following the need for intervention controls
\cite{zhang-nanda-2024-activation}.

\section{Data, Inference, and Evaluation}
\label{sec:setup}
\textbf{Data and models.}
We render XQuAD passages as images, keeping questions in text. The fixed
evaluation set contains 400 aligned items per language: English (EN), Chinese
(ZH), Vietnamese (VI), Thai (TH), and Hindi (HI). All five enter the
behavioral comparison, where Vietnamese (lower-resource, Latin script) and
Chinese (higher-resource, non-Latin) serve as controls; interventions use
English as the donor and Thai and Hindi as targets (EN--TH and EN--HI pairs). A separate 300-item subset estimates held-out steering
directions. These subsets are disjoint by item ID; this does not establish
separation by source passage. ThaiOCRBench comprises Infographics (213 items),
Cognition VQA (217), and Diagram VQA (204): 634 real Thai images resized to a
maximum long side of 1024\,px. It has neither English twins nor oracle
passage transcriptions, so we evaluate only baseline answering and steering
there.

We use Qwen3-VL-8B \cite{Qwen3-VL}, Gemma-3-12B \cite{gemma_2025}, and
InternVL3.5-8B \cite{wang2025internvl3_5}, with 36, 48, and 36 LM blocks.
Inference is frozen, zero-shot, greedy, and bfloat16 on an NVIDIA H100 80GB.
Questions use the target language. Unless otherwise specified, prompts end
with a translated concise-answer suffix:
\emph{Answer directly and concisely without additional explanation.}
Code, item IDs, and per-item outputs:
\url{https://github.com/srn-source/donor-content-controls}.

\textbf{Rendering.} Passages are rendered black on white at 900\,px width with
40\,px padding, a 24\,px font, and 30\,px line height, using one font per
language (DejaVu Sans, Noto Sans SC, Be Vietnam Pro, Garuda, Noto Sans
Devanagari); Thai and Chinese wrap by character, the others by word. Estimated
from image size on Qwen3-VL's 32-px grid, median visual-token counts are about
336, 224, 336, 280, and 252 for EN, ZH, VI, TH, and HI. Fonts, text density,
and token loads are not matched across languages.

\textbf{Behavioral conditions.}
Free-form OCR-VQA supplies the passage image and question. Oracle additionally
supplies the gold passage text, showing how much of the gap closes once the
text itself is available.

\textbf{Scoring.} We score answers with ANLS, which gives partial credit: a prediction scores 1 if it contains the gold answer, and otherwise
one minus its normalized edit distance to the gold, set to zero below 0.5. A
positive score is therefore a partial match, not necessarily an exact answer,
so for key results we also report exact match (ignoring case, whitespace, and
punctuation). Confidence intervals are 95\% bootstrap intervals over items
(10,000 resamples), paired when comparing conditions.

\section{Do Failures Look Like a Language Problem?}
\label{sec:diagnostics}
If output-language behavior contributes to failures, wrong-script
predictions are one observable signal; correct-script predictions, however,
can still be wrong. After establishing the gap, we therefore examine both
output scripts and answer-related signals in hidden states.

\textbf{The gap.}
Table~\ref{tab:behavior} reports free-form and Oracle scores for all five
languages. Free-form scores drop sharply on Thai and Hindi in all three
models, and supplying the passage text (Oracle) recovers most of that drop.
Vietnamese falls much less (0.755--0.820, versus 0.905--0.921 for English),
but Gemma also declines on Chinese (0.602). The large Oracle gains show that
performance improves substantially when the passage text is supplied directly,
although they do not isolate the source of free-form failures.

\begin{table}[tb]
\centering\footnotesize
\caption{Rendered-XQuAD results on 400 aligned items per language. Free-form
and Oracle rows give ANLS with 95\% bootstrap CIs; Free-form supplies the
passage image and question, and Oracle additionally supplies the gold passage
text. Wrong script: share of zero-score free-form outputs written mainly in a
non-target script.}
\label{tab:behavior}
\begin{tabular}{llccccc}
\toprule
Model & Condition & EN & ZH & VI & TH & HI\\
\midrule
Qwen3-VL-8B & Free-form & \ci{.905}{.88}{.93} & \ci{.868}{.84}{.90} & \ci{.820}{.78}{.85} & \ci{.338}{.30}{.38} & \ci{.507}{.46}{.55}\\
 & Oracle & \ci{.916}{.89}{.94} & \ci{.891}{.86}{.92} & \ci{.867}{.84}{.90} & \ci{.851}{.82}{.88} & \ci{.793}{.75}{.83}\\
 & Wrong script (\%) & 0.0 & 6.8 & 0.0 & 8.8 & 1.6\\
Gemma-3-12B & Free-form & \ci{.917}{.89}{.94} & \ci{.602}{.56}{.65} & \ci{.755}{.71}{.79} & \ci{.341}{.30}{.39} & \ci{.622}{.58}{.67}\\
 & Oracle & \ci{.919}{.89}{.94} & \ci{.877}{.85}{.91} & \ci{.867}{.84}{.90} & \ci{.868}{.84}{.90} & \ci{.827}{.79}{.86}\\
 & Wrong script (\%) & 0.0 & 2.8 & 0.0 & 8.9 & 2.9\\
InternVL3.5-8B & Free-form & \ci{.921}{.90}{.95} & \ci{.845}{.81}{.88} & \ci{.774}{.73}{.81} & \ci{.169}{.14}{.20} & \ci{.134}{.10}{.17}\\
 & Oracle & \ci{.944}{.92}{.96} & \ci{.889}{.86}{.92} & \ci{.866}{.83}{.90} & \ci{.813}{.78}{.85} & \ci{.747}{.71}{.79}\\
 & Wrong script (\%) & 0.0 & 1.9 & 0.0 & 14.7 & 0.9\\
\bottomrule
\end{tabular}
\end{table}

\textbf{Are failures in the wrong language?}
The \emph{Wrong script} rows of Table~\ref{tab:behavior} show that only
9--15\% of Thai failures and under 3\% of Hindi failures are written mainly in
a non-target script, usually Latin; most of the rest are written in the target
script but give the wrong answer. These percentages represent the proportion of zero-ANLS predictions written in the wrong script, rather than the proportion of the overall performance gap. Numeric answers and outputs without letters are not classified but
remain in the denominator. A model could, however, hold the answer internally
and still write a wrong one, so we look inside.

\textbf{Is the answer still inside the model?}
For the same failures, LogitLens \cite{nostalgebraist-2020-logit-lens} applies
the final normalization and unembedding to the final input-position state.
A top-10 hit occurs when the target answer's first token ID is among the
ten highest-scoring IDs; a permutation control substitutes another item's
target. At the final layer, Thai hit rates are 0.506, 0.440, and 0.175 for
Qwen, Gemma, and InternVL, against controls of 0.088, 0.040, and 0.016 on 251,
248, and 320 failures; Gemma Hindi yields 0.390 versus 0.044. For Qwen and
InternVL, Devanagari byte splitting makes first-token matches less
informative. This is only a supporting hint, not a bound: a first token in the top ten
does not show that the model holds, or can use, the full answer. Inspecting
states cannot show whether such signals affect the output, so we next intervene
on the state itself.

\section{Does Changing the Language State Recover Answers?}
\label{sec:interventions}
\subsection{Patching in an English state}
\label{sec:patch}
If language commitment contributes to a failure, replacing the state at
the final input position with an English donor state may change the output
language and let the model produce an English answer. For each English--target pair, we cache the English run's hidden
states and, during the target run's prefill, replace the output of the final
transformer block (35 for Qwen/InternVL, 47 for Gemma) at the final input
position with the English one ($P$); generated tokens are not patched, and
$P_0$ is the unpatched baseline. We patch only the final position, since
sequence lengths differ across languages; other positions may also matter.

We measure two things. \emph{Script composition} is the fraction of an
output's non-whitespace characters in the target script or in Latin, averaged
over outputs.
\emph{EN-gold recovery} is the fraction of baseline failures (target-gold
ANLS${}=0$ in the unpatched run) whose output now matches the item's English
gold (ANLS${}>0$).

With aligned donors ($P$ rows of Table~\ref{tab:donor}), most output switches
to Latin script (Latin fraction .620--.820), and many baseline failures now
match the English gold (.132--.447). At first glance, these recoveries suggest that the recipient already
contained the answer, but the donor itself may carry answer information.

\begin{table}[tb]
\centering\footnotesize
\caption{Final-block patching with paired ($P$) and shuffled ($C$) English
donors. Latin uses all 400 items; other columns are fractions of baseline
failures scoring positively (columns can overlap) against the recipient's
English gold (95\% CI; Exact: exact match), the shuffled donor's English
answer (exact match in parentheses), or the target-language gold.}
\label{tab:donor}
\begin{tabular}{llrrrrr}
\toprule
Model & Patch & Latin & EN-gold rec. & Exact & Donor match & Target rec.\\
\midrule
\multicolumn{7}{l}{\emph{EN$\to$TH}}\\
Qwen & $P$ & .762 & .430 [.371,.494] & .243 & -- & .048\\
 & $C$ & .607 & .044 [.020,.072] & .000 & .161 (.020) & .032\\
Gemma & $P$ & .675 & .233 [.181,.289] & .076 & -- & .056\\
 & $C$ & .624 & .048 [.024,.076] & .000 & .041 (.004) & .016\\
InternVL & $P$ & .820 & .150 [.113,.191] & .059 & -- & .034\\
 & $C$ & .825 & .053 [.031,.078] & .000 & .019 (.000) & .016\\
\midrule
\multicolumn{7}{l}{\emph{EN$\to$HI}}\\
Qwen & $P$ & .708 & .447 [.378,.516] & .255 & -- & .037\\
 & $C$ & .535 & .016 [.000,.037] & .000 & .153 (.005) & .020\\
Gemma & $P$ & .620 & .257 [.184,.331] & .103 & -- & .044\\
 & $C$ & .520 & .074 [.037,.118] & .008 & .060 (.000) & .075\\
InternVL & $P$ & .771 & .132 [.096,.167] & .041 & -- & .056\\
 & $C$ & .783 & .064 [.041,.091] & .000 & .023 (.000) & .047\\
\bottomrule
\end{tabular}
\end{table}

\subsection{Whose answer is it?}
\label{sec:shuffle}
The aligned donor, however, comes from the same question in English, so its
state may carry the answer itself. We therefore repeat the patch with a
shuffled donor ($C$): the English state of a different item, assigned by a
fixed derangement. A different item usually, though not always, has a
different answer.

With shuffled donors the Latin fraction stays high (.520--.825), but EN-gold
recovery falls to .016--.074. Exact matching sharpens the contrast: .041--.255
for aligned donors versus .000--.008 for shuffled ones. Script changes remain substantial when donor identity and content change,
whereas recipient EN-gold recovery drops sharply, so the two effects depend
differently on donor pairing. Shuffled-donor outputs often partially match the shuffled donor's own
answer (.019--.161) but rarely match it exactly (.000--.020), a pattern consistent
with partial donor-content transfer, although some matches may reflect
incidental string overlap. Target-gold
recovery stays low under both donors (.016--.075): changing the output
language rarely repairs the target-language answer. Patching alone therefore
cannot show that the model held the answer, because the donor can supply it.

\subsection{Changing the language without a single donor}
\label{sec:steer}
Shuffled-donor patching suggests that replacing a state can transfer answer
content along with language. To reduce dependence on the content of any single
donor, we use
mean-difference activation steering
\cite{turner2023activation,rimsky-etal-2024-steering}. The direction is the
mean English minus the mean target-language final-position state; because it averages
over many items, it is tied to no single question, though averaging need not
remove all content. $S_+$ adds it to
the final input position, scaled by $\alpha$, and $S_-$ subtracts it.
$S_{\mathrm{rand}}$ adds a seeded random direction of the same norm, which
controls for disturbing the state at all. We steer at about 75\% depth (block
27 for Qwen/InternVL, 36 for Gemma) with $\alpha=1$. This setting was chosen
exploratorily on the evaluation items and not validated on separate data. To test whether the direction itself transfers,
we also fit it on 300 disjoint items.

Table~\ref{tab:steer} reports steering at 75\% depth. With in-sample
directions, $S_+$ raises the Latin fraction to .386--.660, against
.019--.146 under the random control, and EN-gold recovery to .053--.137,
against .000--.056. Because the random control also recovers some answers,
we report the excess over it, using directions fitted on held-out items:
.045--.083, with all six intervals above zero. Under exact match the excess
shrinks to .010--.060; five of six intervals exclude zero, and Qwen Hindi's
lower bound is zero.

The random control leaves script composition nearly unchanged but is not
perfectly inert for content: its ANLS change relative to the unpatched
baseline ranges from $-$.018 to $+$.023, and only Qwen Hindi and InternVL
Hindi have intervals excluding zero, in opposite directions. These intervals
cover items, not random seeds: each model uses one seeded random direction.
Failure to recover with this direction does not rule out recovery by other
interventions. Target-directed $S_-$ increases target-script composition
without improving aggregate ANLS. The steering excess is modest, and it does
not show that every steering recovery reflects the recipient's own knowledge.

\begin{table}[tb]
\centering\footnotesize\setlength{\tabcolsep}{4pt}
\caption{Steering at 75\% depth, $\alpha=1$. Latin and EN-gold columns use
in-sample directions. Excess: EN-gold recovery of $S_+$ minus $S_{\mathrm{rand}}$
with held-out directions, under ANLS and exact matching. $\Delta$ANLS: ANLS
change under $S_{\mathrm{rand}}$ versus the unpatched baseline. Brackets:
paired 95\% CIs.}
\label{tab:steer}
\begin{tabular}{llccccccc}
\toprule
 & & \multicolumn{2}{c}{Latin} & \multicolumn{2}{c}{EN-gold rec.}
 & \multicolumn{2}{c}{Excess} & \\
\cmidrule(lr){3-4}\cmidrule(lr){5-6}\cmidrule(lr){7-8}
Model & Lang. & $S_+$ & $S_{\mathrm{rand}}$ & $S_+$ & $S_{\mathrm{rand}}$
 & ANLS & Exact & $\Delta$ANLS $S_{\mathrm{rand}}$\\
\midrule
Qwen & TH & .487 & .107 & .088 & .036 & \ci{+.052}{.028}{.080} & \ci{+.024}{.008}{.044} & \ci{$-$.010}{$-$.024}{+.002}\\
 & HI & .393 & .028 & .053 & .005 & \ci{+.045}{.020}{.074} & \ci{+.010}{.000}{.025} & \ci{$-$.018}{$-$.035}{$-$.001}\\
Gemma & TH & .386 & .107 & .084 & .028 & \ci{+.057}{.029}{.090} & \ci{+.016}{.004}{.033} & \ci{+.011}{$-$.007}{+.030}\\
 & HI & .413 & .020 & .081 & .000 & \ci{+.083}{.038}{.128} & \ci{+.060}{.023}{.105} & \ci{$-$.009}{$-$.024}{+.006}\\
InternVL & TH & .660 & .146 & .109 & .031 & \ci{+.078}{.050}{.109} & \ci{+.028}{.009}{.050} & \ci{$-$.005}{$-$.017}{+.006}\\
 & HI & .559 & .019 & .137 & .056 & \ci{+.076}{.050}{.105} & \ci{+.038}{.020}{.058} & \ci{+.023}{+.007}{+.040}\\
\bottomrule
\end{tabular}
\end{table}

\subsection{Does it hold on real images?}
\label{sec:real}
All results so far use rendered passages. ThaiOCRBench free-form ANLS is 0.521, 0.414, and 0.239 for Qwen, Gemma,
and InternVL. Applying the held-out rendered-XQuAD direction unchanged
raises Latin fractions from 0.10--0.16 to 0.24--0.48 at $\alpha=1$,
while the sampled random control remains near baseline script composition.
Target-directed steering leaves ANLS flat or lower, so script control
transfers across domains without an answer-quality benefit.

\section{Discussion and Limitations}
Our results do not support reading most paired-patching recovery as evidence
that the recipient already knew the answer, but they do not bound how many
failures involve language commitment. Rendering
is not matched across languages, so visual-encoding difficulty may contribute
to the gap. Steering intervals do not cover variation across random directions or the
exploratory choice of layer and strength, so they are not fully
confirmatory, and
a position control that writes the same final vector is needed for stronger
localization claims. Results also vary across models: InternVL is weakest
on free-form Thai and Hindi, LogitLens, and aligned patching, yet has a
comparable steering excess. Three
8--12B models with Thai-only real-image transfer do
not establish generality across scales.

\textbf{Relevance to developing regions.} Thai and Hindi are read by hundreds
of millions of people, yet remain weak points for open VLMs. For teams building document
systems in these settings, the practical lesson is that improving script
adherence does not necessarily improve answer accuracy: evaluations should report
script adherence and answer correctness separately.

\section{Conclusion}
Multilingual OCR-VQA models often fail on Thai and Hindi images. One possible
explanation is that they hold answer information but fail to express it in the
target language; we asked whether hidden-state interventions recover previously
failed answers and how donor content affects that recovery. Cross-lingual
activation patching changes the output script and produces English-gold
matches, but these apparent recoveries depend strongly on donor content:
shuffled donors preserve substantial script changes while sharply reducing
matches, and mean-difference steering yields only modest recovery beyond random
perturbations. Script change, recipient-answer matching, and
donor-content dependence should therefore be evaluated separately. Changing how a
model expresses an answer does not necessarily mean recovering what the model
already knows.
\label{maintextend}
\clearpage
\bibliographystyle{splncs04}
\bibliography{main}
\end{document}